\documentclass[letterpaper, 10 pt, conference, twoside]{IEEEtran}

\IEEEoverridecommandlockouts                              

\makeatletter
\let\NAT@parse\undefined
\makeatother

\usepackage[T1]{fontenc}
\usepackage[hidelinks]{hyperref}
\usepackage{lineno}
\usepackage{graphicx}
\usepackage{subfigure}
\usepackage[nolist,nohyperlinks]{acronym}
\usepackage{amsmath}
\usepackage{float}
\usepackage{epstopdf}
\usepackage{url}
\usepackage{graphicx}
\usepackage{transparent}
\usepackage{pgfplots} 
\pgfplotsset{compat=newest}
\usetikzlibrary{plotmarks}
\usepackage{grffile}
\usepackage{amsmath}

\newlength\figureheight 
\newlength\figurewidth 

\graphicspath{{images/}}

\newcommand*{\matlabtikz}{matlab_plots/}
\definecolor{green}{RGB}{0,150,0}
\definecolor{darkblue}{rgb}{0,0,0.5} 
\definecolor{fuchsia}{RGB}{127, 0, 127}
\definecolor{darkpurple}{RGB}{255, 0, 255}
\definecolor{bluegreen}{RGB}{0,127,127}
\definecolor{lightblue}{RGB}{0,127,255}
\definecolor{lightpurple}{RGB}{127,127,255}

\newacro{NMXM}[NMXM]{Normalized Maximum Cross-Correlation Magnitude }
\newacro{MSE}{Mean Squared Error}
\newacro{VAR}{Variance}
\newacro{STD}{Standard Deviation}
\newacro{MAVs}{Micro Aerial Vehicles}
\newacro{MAV}{Micro Aerial Vehicle}
\newacro{FOV}{Field of View}

\title{\LARGE \bf
Efficient Optical Flow and Stereo Vision for Velocity Estimation and Obstacle Avoidance 
on an Autonomous Pocket Drone 
}
\author{Kimberly McGuire$^{1}$, Guido de Croon$^{1}$, Christophe De Wagter$^{1}$, Karl Tuyls$^{1,2}$ and Hilbert  Kappen$^{3}$
	\thanks{$^{1}$ Delft University of Technology, The Netherlands 
		\newline
	{\tt\small k.n.mcguire@tudelft.nl}, \newline {\tt\small g.c.h.e.decroon@tudelft.nl}}
\thanks{$^{2}$ University of Liverpool, United Kingdom ~~~ }
\thanks{$^{3}$ Radboud University of Nijmegen, The Netherlands}
}
\begin{document}
	
\maketitle

\begin{abstract}
\ac{MAVs} are very suitable for flying in indoor environments, but autonomous navigation is challenging due to their strict hardware limitations. This paper presents a highly efficient computer vision algorithm called Edge-FS for the determination of velocity and depth. It runs at 20 Hz on a 4~g stereo camera with an embedded STM32F4 microprocessor (168 MHz, 192 kB) and uses edge distributions to calculate optical flow and stereo disparity. The stereo-based distance estimates are used to scale the optical flow in order to retrieve the drone's velocity. The velocity and depth measurements are used for fully autonomous flight of a 40~g pocket drone only relying on on-board sensors. This method allows the MAV to control its velocity and avoid obstacles.
\end{abstract}


\section{Introduction}

\begin{figure}[t]
	\centering

	\def\svgwidth{0.8\linewidth}
	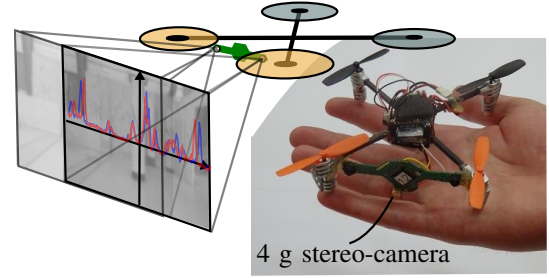
	\vspace{-0.3cm}
	\caption{Pocket drone with a lightweight forward looking 4g stereo camera. A very efficient vision algorithm runs embedded on the STM32F4 processor (168 MHz, 192 kB), to determine velocity and depth necessary for the pocket drone's visual navigation. }
	\label{fig:front_image}
\end{figure}

\IEEEPARstart{D}{eployment} of Micro Aerial Vehicles (MAVs) is important for indoor tasks such as inspections, search-and-rescue operations, green house observations and more. Tiny MAVs, also called pocket drones (\textless  50~g, as in Fig.~\ref{fig:front_image}), are ideal for maneuvering through very narrow spaces, as often occurs in indoor environments. In order for them to autonomously navigate through a GPS-deprived area, there are several on-board sensors to consider (laser rangers, motion sensors, infrared rangers, sonar). The pocket drone's sensor of choice is a RGB camera. It is the most energy efficient and versatile sensing option,
as multiple variables can be observed from the image stream: obstacles, motion, object recognition and more.

Using cameras enables the \ac{MAV} to extract essential information for autonomous navigation. A stereo vision setup with two cameras has been particularly successful, for instance for obstacle avoidance~\cite{hu2010evaluation}. Since there are strict limitations on energy expenditure, sensing, and processing capabilities on a pocket drone, even relatively efficient stereo vision methods~\cite{hirschmuller2008stereo}\cite{geiger2011stereoscan} are computationally too heavy to run on-board a microprocessor. Therefore, an even more efficient stereo vision algorithm was developed, which is able to run at 10~Hz on a 20~g flapping wing \ac{MAV}, the DelFly Explorer~\cite{de2014autonomous}. It is still the lightest fully autonomous \ac{MAV} to this date, which can fly through a room and avoid obstacles with purely onboard sensing and processing~\cite{tijmons2016obstacle}. 

Since tailed flapping wing \ac{MAV}s such as the DelFly Explorer are passively stable, there is no need to compute their velocity to compensate for drift. However, for inherently unstable platforms like a quadcopter, velocity estimation is necessary for stabilization when navigating in constrained areas.
Optical flow is the way in which objects move in two sequential images and is the most important visual cue for velocity estimation.  It can be calculated in a dense manner (Horn-Schunck~\cite{Horn1981}, F\"arneback~\cite{farneback2003two}) or a sparse manner, e.g., by tracking features such as  Shi-Tomasi~\cite{Shi} or FAST~\cite{Rosten2005} over time with a Lucas-Kanade tracker~\cite{bouguet2001pyramidal}. These types of techniques have proven themselves on numerous occasions~\cite{honegger2013open}, nonetheless do require a platform with a decent amount of computing power. On a pocket drone such standard optical flow methods either cannot be run in real-time or take  consist of an unpractically large part of the processing time, leaving little to no room for other types of processing.
Especially when autonomous flight is the final goal, optical flow determination will only constitute a part of what the \ac{MAV} has to do, as much more information can be retrieved from the image stream.

 In order to design a computationally much more efficient optical flow algorithm, we have drawn inspiration from the study in~\cite{Lee2004}, which proposed using spatial edge distributions to track motion in the image. Specifically, in~\cite{mcguire2015}, we presented EdgeFlow, which improved upon the work in~\cite{Lee2004} by introducing a variable time horizon for determining sub-pixel flow. EdgeFlow ran embedded at 30~Hz on a lightweight stereo camera positioned underneath a pocket drone.  The stereo camera was pointing down and was estimating optical flow and a global height estimate, assuming that it was looking at a flat ground surface. With these, the \ac{MAV} determined its own velocity and used this in a guided control, where it autonomously matched externally-given velocity references. However, a 4~g stereo camera for a 40~g pocket drone is significant, so it is a waste to have this   ``heavy'' sensor looking downward and  not using it to avoid obstacles in the flight direction.

This paper presents a major extension of EdgeFlow, which enables the stereo camera to face forward on a \ac{MAV}, so it can be used for navigation purposes. As the pocket drone will now be facing hallways, rooms, doors etc., the assumption of looking straight at a flat plane will not hold anymore.  The same matching paradigm used to determine EdgeFlow, will now be used to not only calculate optical flow but also stereo depth over the  entire image. \emph{EdgeStereo}, as called for convenience, uses the so-determined distances to properly scale the locally observed optical flow in order to retrieve a velocity estimate. This combination of EdgeFlow and EdgeStereo will be called \emph{Edge-FS}.

Our main contribution is that the presented method provides both velocity and distance estimates, while still being computationally efficient enough to run close to the frame rate on a very limited embedded processor. As such, the method enables unstable MAVs such as tiny quadcopters to perform fully autonomous flights in unknown environments. The EdgeFlow and EdgeStereo methods will be explained in more detail in section~\ref{sec:edgeflow}. Off-line results for velocity estimates with a set of images is shown in section~\ref{sec:simulation}. From here, the algorithm is embedded on the lightweight stereo camera and placed on 40~g pocket drone for velocity estimation (section~\ref{sec:velocityest}). Finally, the velocity estimate is used together with EdgeStereo-based obstacle detection to perform fully autonomous navigation in an environment with obstacles (section~\ref{sec:closedloopflight}). This is followed by some concluding remarks.

\subsection{Related Work}\label{sec:relatedwork}

In related research, several works have achieved optical flow based control of a \ac{MAV}, e.g.,~\cite{Grabe2015}\cite{Romero2009}\cite{Kendoul2009}. As mentioned in the introduction, the standard optical flow methods are computationally too heavy to run on a quadcopter of less than 50~g. For instance, Dunkley et al. have flown with a 25~g quadcopter before, while computing optical flow for visual odometry~\cite{dunkley2014visual}. However, this was done on an external computer. As miniaturization of hardware also poses a limitation on communication bandwidth, this can result in a significant delay in the controls. To obtain full autonomy, it would be wise to uncouple a \ac{MAV} of any external dependencies.

 To design extremely lightweight \ac{MAV}s for autonomous flight, some researchers looked into EMD sensors~\cite{Ruffier2003} and  other 1D signal sensors~\cite{green2008optic}. Briod et al.~\cite{Briod2013} proposed the design of a 45~g quadcopter for optical flow based control with 1D  flow sensors. They followed up with this research on a heavier 278~g platform containing 8 of these sensors  pointing in all directions~\cite{briod2016method}. With this they could hover the quadcopter in various cluttered environments. The results are impressive, nevertheless they were achieved by using multiple single purpose sensors. As they can only sense motion, it does not leave much room to detect other variables necessary for navigation.

More similar to our research, Moore et al. implemented an efficient optic flow algorithm on a small lightweight (2 g) omnidirectional camera system on a 30~g helicopter~\cite{Moore}. With a ring of 8 low-resolution image chips (64 x 64 pixels), the \ac{MAV} could compute optical flow. It did this by computing the edges, compressing the images and calculate the displacement by block matching which resulted in translational optical flow. The vision calculations where done on-board the helicopter with 10 Hz, yet the flight controls where computed off-board. Although the potential of a full on-board implementation is there, the redundancy lies in the ratio of
cameras to sensed variables. One camera has the potential of detecting flow in 3 directions; they used 8 to only detect 2 (forward and sideways velocity).

Optical flow can also be used to detect obstacles~\cite{mori2013first}, however the \ac{MAV} needs to be constantly on the move. This is not required if stereo vision is used for depth information. With this, Oleynikova et al. developed a reactive avoidance controller for a quadcopter (30 cm in diameter)~\cite{oleynikova2015reactive}.  From the obtained stereo disparity map, they accumulated the values along the columns to get a summed disparity factor. Assuming that the obstacles are vertical and long, these can be detected quickly.  The stereo map was calculated over the entire image first before accumulation to a vector. This significantly impacts the amount of computation making it less suitable for implementation on a smaller \ac{MAV}.

\section{Velocity and Depth from Edges}\label{sec:edgeflow}

To achieve autonomous navigation with a camera on an unstable pocket drone, we need to obtain two variables: velocity and depth. In the introduction we mentioned that many of the mainstream computer vision will be computationally too heavy to run on the pocket drone. In~\cite{mcguire2015}, we presented EdgeFlow, which can detect optical flow within the image in a semi-dense but computationally efficient manner, embedded on a 4~g stereo board. During the experiments, the stereo camera looked down to the ground, estimating the pocket drone's forward and sideways velocity. This section will explain the modifications that are necessary to make the stereo camera point forward and still be able to measure those variables.  EdgeFlow will be concisely recapped. Subsequently, we will present its extension with EdgeStereo to Edge-FS, which will be used for obstacle detection in the experiment part of this paper.

\subsection{From Camera to State}

\begin{figure}[t]
	\centering
	\vspace{0.2cm}
	\def\svgwidth{0.8\linewidth}
	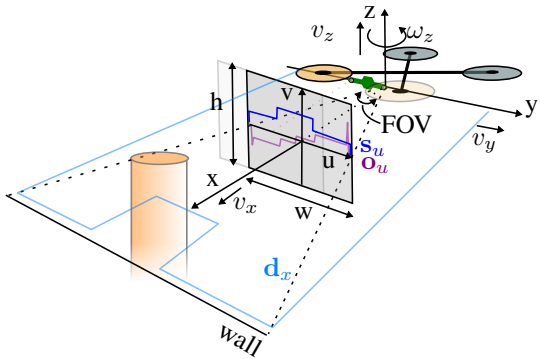
	\vspace{-0.2cm}
	
	\caption{The MAV's body fixed coordinates with respects to the camera axis, shown for the left camera (XYZ). The conventional aircraft coordinates of east north up is used for the \ac{MAV} as the camera. The image coordinates in width and height are represented as $u$ and $v$ respectively.}
	\label{fig:camera_axis}
\end{figure}

When looking orthogonally at a planar ground surface while moving, the optical flow field is rather simple and allows for easy determination of the forward and sideways velocities with the help of a single height measurement. But to navigate without bumping into anything, the \ac{MAV} needs to see objects
in the direction of motion, which in this study is forward. Due to the likely non-planar (3D structure) of the environment in forward direction, the optical flow field will become more complex. Moreover, the forward velocity now can only be observed by means of the divergence of optical flow, which is more difficult to determine, especially close to the focus of expansion. Here we delve into how we determine the velocities with the help of  forward facing stereo images. In principle, the unscaled velocities and the rotation rates can be determined from the image alone, according to the paper of~\cite{longuet1980}. Longuet-Higgins and Prazdny implied that measured flow ($\mathbf{o}_u$) is the summation of a translational flow ($\mathbf{o}^T_u$) and rotational component ($\mathbf{o}^R_u$). Before estimating the horizontal planar velocity, we first have to determine $\mathbf{o}^R_u$.

Although~\cite{longuet1980} assumes rotations in all directions, we can make simpler assumptions for the pocket drone. Fig.~\ref{fig:camera_axis} shows the placement and axis definition of the drone and camera. For obstacle avoidance it is essential to look in the direction of motion, which in this case is the direction of the positive $x$ axis. Here, correctional pitch and roll motion for drift compensation will be relatively small, but yaw rotations will be more common. Assuming that the latter only has significant effect on the optical flow, $\mathbf{o}^R_u$ can be approximated (assuming small angles)  using the gyroscopes on the on-board IMU of the pocket drone: 
\vspace{-0.3cm}
\begin{eqnarray}
 {o}^R_{u,i} \approx \omega_Z \cdot \dfrac{w}{\alpha_{FOV}} \\ \label{eq:derotation}
 \mathbf{o}^R_u = [ {o}^R_{u,1} , \dots, {o}^R_{u,w}]
\end{eqnarray}
where $w$ is the width of the image, $\alpha_{FOV}$ is the angle of the Field of View (FOV) and  $\omega_Z$ is the yaw rotation measured from the gyroscopes.

Now that $\mathbf{o}^R_u$ is known, we can  isolate  $\mathbf{o}^T_u$ to determine the pocket drone's forward ($v_x$) and sideways velocity. With the coordinate system we use in this paper (Fig.~\ref{fig:camera_axis}), Longuet's equation of $\mathbf{o}^T_u$ is expressed as:

\vspace{-0.3cm}
\begin{eqnarray}
\mathbf{o}^T_u = (-v_y + \mathbf{x} v_x)/\mathbf{d}_x \\
\mathbf{d}_x \mathbf{o}^T_u = -v_y + \mathbf{x} v_x \label{eq:linefit} 
\end{eqnarray}

Where $\mathbf{x}$ is an array of indices of the image columns.  Depth, $\mathbf{d}_x $, scales the optical flow resulting in motion parallax, as close objects appear to move faster than objects far away.  In~\cite{mcguire2015}, a global height estimate was used to scale the optical flow back to velocities, which is sufficient if the camera is looking at a flat floor or perpendicular to a straight wall. This assumption will not hold when the \ac{MAV} is flying towards a wall at an angle or whenever obstacles at different distances are in the field of view. This non-constant depth needs to be accounted for when scaling the optical flow, therefore the stereo depth is needed over the entire size of the image for a better velocity estimate. Local right-left image disparity from a stereo camera can be transformed to actual depth in meters by using the camera parameters, with the following approximation:
 \vspace{-0.2cm}
\begin{eqnarray}\label{eq:depth}
\mathbf{d}_x \approx \dfrac{w\cdot r }{ \alpha_{FOV} \cdot \mathbf{s}_u}~ 
\end{eqnarray}

 \noindent where $r$ is the baseline between the two cameras, and the stereo disparities in pixels along the image columns  is $\mathbf{s}_u$.

With depth $\mathbf{d}_x$ and translational optical flow $\mathbf{o}^T_u$, it is now possible to calculate the \ac{MAV}'s sideways and forward velocity by fitting a linear model to \eqref{eq:linefit}. 
In the next section we will explain how to obtain both optical flow and stereo depth from a stream of low resolution stereo images.

\subsection{Procedure for Edge-FS}\label{sec:method}

\begin{figure}[!t]
	\vspace{0.2cm}
	\centering
	\scriptsize
	\def\svgwidth{1.0\linewidth}
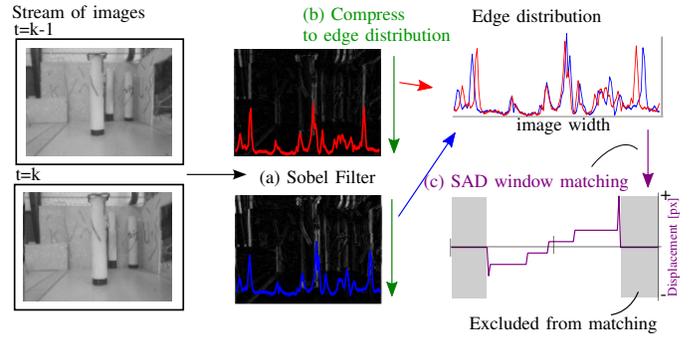

		\vspace{-0.3cm}
		
	\caption{
		The matching algorithm for both EdgeFlow as EdgeStereo. The images' gradients  (a) calculated by a Sobel filter, (b) summed up to  an edge distribution. These are (c) matched with other edge distribution.  The gray areas are excluded sections (equal to the range plus half SAD block size).
	}
	\label{fig:image_pipeline}
\end{figure}

\begin{figure}[!t]
	\centering
		\scriptsize
	\def\svgwidth{\linewidth}
	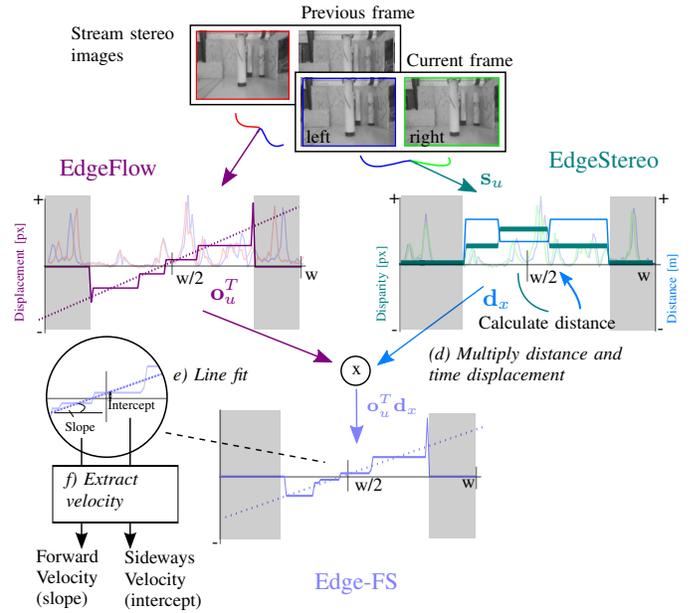
	\vspace{-0.3cm}
	\caption{ The temporal pixel disparities per column of EdgeFlow is (d) scaled by EdgeStereo. Following (\ref{eq:linefit}), e) a line fit is done on this array of values, from which the forward and sideways velocities can be extracted from the slope and intercept, respectively. \vspace{-0.3cm}
	}
	\label{fig:image_pipeline_linefit}
\end{figure} 
%

The matching principle of EdgeFlow is shown in Fig.~\ref{fig:image_pipeline}. The images A and B are first filtered with a Sobel filter to get the horizontal gradients (Fig.~\ref{fig:image_pipeline}(a)). These gradients are summed along the rows (compressed) to a (spatial) edge  distribution (Fig.~\ref{fig:image_pipeline}(b)). From image A and B, the edge distributions are compared with Sum of Absolute Differences (SAD) block matching (Fig.~\ref{fig:image_pipeline}(c)). This locates the similar patches of the edge distributions within a certain distance from each other, to obtain the pixel displacement between the two. 

If image A and B from Fig.~\ref{fig:image_pipeline} are two temporal sequential images ($t$ with $t-1$) in time, this will result in the pixel flow, thus EdgeFlow as presented in~\cite{mcguire2015}.  Based on the previous flow value, EdgeFlow adaptively chooses how far in time ($t-n$) it will compare the current edge distribution to. On top of that, the flow shift predicted by a yaw rotation ($\mathbf{o}^R_u$, as calculated in (\ref{eq:derotation}) will shift the start of the block matching scheme.  This and the adaptive time horizon will be present for the experiments in this paper. Note that in~\cite{mcguire2015}, also the direction along the image height was used to estimate the forward velocity (as the camera was looking down). In this paper, it will not be used as the forward velocity ($v_x$) will now be subtracted from the divergence of Edge-FS.

 Previously in~\cite{mcguire2015}, the entire edge distribution of the left and right image were matched to obtain a global depth estimate. To get a better velocity estimate with a forward camera, we need to use pixel disparity per column. To calculate both column-wise optical flow and stereo vision and keep the algorithm computationally efficient, the exact same matching principle of EdgeFlow (Fig.~\ref{fig:image_pipeline}) is used, resulting in EdgeStereo. Disparity to depth in meters is calculated with the known camera parameters and (\ref{eq:depth}) from the last section. Sequentially, EdgeStereo scales EdgeFlow to  compensate for the motion parallax (see Fig.~\ref{fig:image_pipeline_linefit}(d)), which results in the left side of (\ref{eq:linefit}). These values will then be fitted to a linear model (Fig.~\ref{fig:image_pipeline_linefit}(e)), which gives us the slope and intercept of the line. With the camera parameters, the forward and sideways velocities are estimated (Fig.~\ref{fig:image_pipeline_linefit}(f)).

\section{Off-line Vision Experiments}\label{sec:simulation}

\begin{figure}[t]
	\vspace{3pt}
	\addtolength{\subfigcapskip}{+10pt}
		\small
	\centering
	\subfigure[\label{subfig:database_forward}]{
		
		\includegraphics[width=0.32\linewidth]{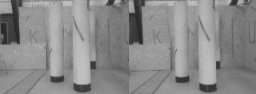} \hfill \includegraphics[width=0.32\linewidth]{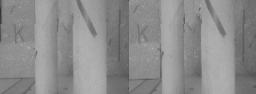} \hfill
		\includegraphics[width=0.32\linewidth]{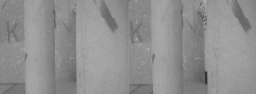}

}
	\addtolength{\subfigcapskip}{-10pt}
	\vspace{-0.3cm}
	
	\centering
		\small 
\subfigure[\hspace{7cm}\label{fig:matlab_simulation_forward}]{

	\setlength\figureheight{5cm} 
\setlength\figurewidth{0.8\linewidth}
\input{\matlabtikz/Edgeflow_Farneback_board_1_data_4.tex}
\vspace{-0.3cm}
}

	\caption{(a) 	Several screen shots of the set of images used for off-line estimation of the velocity. Here the diversity in amount of texture can be seen. (b) Off-line velocity estimate calculated by Edge-FS and F\"arneback, held against the ground truth for the forward moving camera's data set. } 
	\label{fig:matlab_simulation}
\end{figure}

Before implementing the algorithm on the actual stereo board, EdgeFlow was run on a set of stereo-images in MATLAB
(version R2015b on a Dell Latitude E7450, i7-5600U CPU @ 2.60GHz processor). Fig.~\ref{subfig:database_forward} shows screen shots of the data set used in this section, where the camera moves towards obstacles at different distances. In  Fig.~\ref{fig:matlab_simulation_forward},  EdgeFlow scaled by EdgeStereo, now dubbed as Edge-FS, results in the velocity estimates. 

Edge-FS is contrasted against the well-known optical flow method developed by F\"arneback~\cite{farneback2003two}, a dense optical flow method (Fig.~\ref{fig:matlab_simulation}). Although less used than a more conventional KLT-tracker~\cite{bouguet2001pyramidal}, preliminary analyses indicated it to be more suited for the low-resolution images used here. With its default parameters set as in MATLAB R2015b, the sparse magenta line illustrates that the KLT-tracker indeed has difficulties with the low-quality, low-resolution images (128 x 96 pixels).

For F\"arneback, depth is determined by matching the stereo-images with each other  and converting the resulting pixel disparity to a distance. To get velocity, the same line-fit is used as for EdgeFlow\footnote{The EdgeFlow code as embedded on the stereo camera has a mean computation time for EdgeFlow is 0.00134 seconds (compiled for Linux) and for F\"arneback is 0.00466 seconds  on the same stereo-image data set.}, but here the whole image is considered rather than the compressed form like the edge distributions. After comparison of the methods with different parameters, both Edge-FS and F\"arneback are set up with a window size of 11 pixels and a search range of 15 pixels (F\"arneback's pyramid level at 1).  Both forward (x) and sideways (y) velocity measurements, shown in Fig.~\ref{fig:matlab_simulation}, are compared against the ``ground truth'' as obtained with an OptiTrack motion capture system\footnote{\url{www.optitrack.com}}, with 24 infrared cameras. The plots also include several values to determine the quality of the velocity estimates: \ac{MSE},  \ac{VAR} and \ac{NMXM}.   A low \ac{MSE} indicates greater similarity and low \ac{VAR} is a smaller spread of the measurement from the ground truth.  A high \ac{NMXM} stands for a better shape correlation between the two. All these metrics indicate Edge-FS to obtain more accurate results on this data set than the computationally more expensive F\"arneback method.

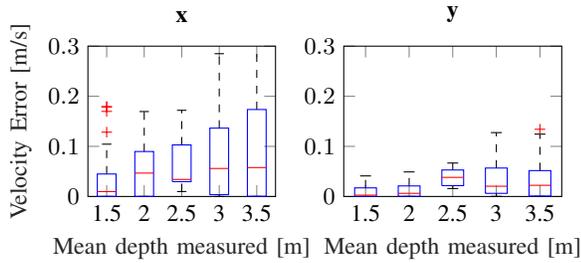
\begin{figure}[t]
	\centering
		\small
			\setlength\figureheight{2cm} 
	\setlength\figurewidth{0.8\linewidth}
%
%
\begin{tikzpicture}

\begin{axis}[%
width=0.352\figurewidth,
height=\figureheight,
at={(0\figurewidth,0\figureheight)},
scale only axis,
xmin=1.5,
xmax=6.5,
xtick={1,2,3,4,5,6},
xticklabels={{0},{1.5},{2},{2.5},{3},{3.5}},
xlabel style={font=\color{white!15!black}},
xlabel={Mean depth measured [m]},
ymin=0,
ymax=0.3,
ylabel style={font=\color{white!15!black}},
ylabel={Velocity Error [m/s]},
axis background/.style={fill=white},
title style={font=\bfseries},
title={x},
title style={font={\small\bfseries}},legend style={font=\tiny},
]
\addplot [color=black, dashed, forget plot]
  table[row sep=crcr]{%
2	0.0448810190709668\\
2	0.104377335547685\\
};
\addplot [color=black, dashed, forget plot]
  table[row sep=crcr]{%
3	0.0897542941407297\\
3	0.169524904472501\\
};
\addplot [color=black, dashed, forget plot]
  table[row sep=crcr]{%
4	0.102885040933049\\
4	0.17201233123911\\
};
\addplot [color=black, dashed, forget plot]
  table[row sep=crcr]{%
5	0.136397085417089\\
5	0.284948344961733\\
};
\addplot [color=black, dashed, forget plot]
  table[row sep=crcr]{%
6	0.173584241220238\\
6	0.33\\
};
\addplot [color=black, dashed, forget plot]
  table[row sep=crcr]{%
2	1.25950535463382e-05\\
2	0.000176368734398302\\
};
\addplot [color=black, dashed, forget plot]
  table[row sep=crcr]{%
3	6.17734350250032e-05\\
3	0.000180742663897782\\
};
\addplot [color=black, dashed, forget plot]
  table[row sep=crcr]{%
4	0.00988029103908339\\
4	0.029565185294512\\
};
\addplot [color=black, dashed, forget plot]
  table[row sep=crcr]{%
5	1.11728782643539e-05\\
5	0.00354994750730153\\
};
\addplot [color=black, dashed, forget plot]
  table[row sep=crcr]{%
6	3.10138046000219e-06\\
6	0.000399291080679376\\
};
\addplot [color=black, forget plot]
  table[row sep=crcr]{%
1.875	0.104377335547685\\
2.125	0.104377335547685\\
};
\addplot [color=black, forget plot]
  table[row sep=crcr]{%
2.875	0.169524904472501\\
3.125	0.169524904472501\\
};
\addplot [color=black, forget plot]
  table[row sep=crcr]{%
3.875	0.17201233123911\\
4.125	0.17201233123911\\
};
\addplot [color=black, forget plot]
  table[row sep=crcr]{%
4.875	0.284948344961733\\
5.125	0.284948344961733\\
};
\addplot [color=black, forget plot]
  table[row sep=crcr]{%
1.875	1.25950535463382e-05\\
2.125	1.25950535463382e-05\\
};
\addplot [color=black, forget plot]
  table[row sep=crcr]{%
2.875	6.17734350250032e-05\\
3.125	6.17734350250032e-05\\
};
\addplot [color=black, forget plot]
  table[row sep=crcr]{%
3.875	0.00988029103908339\\
4.125	0.00988029103908339\\
};
\addplot [color=black, forget plot]
  table[row sep=crcr]{%
4.875	1.11728782643539e-05\\
5.125	1.11728782643539e-05\\
};
\addplot [color=black, forget plot]
  table[row sep=crcr]{%
5.875	3.10138046000219e-06\\
6.125	3.10138046000219e-06\\
};
\addplot [color=blue, forget plot]
  table[row sep=crcr]{%
1.75	0.000176368734398302\\
1.75	0.0448810190709668\\
2.25	0.0448810190709668\\
2.25	0.000176368734398302\\
1.75	0.000176368734398302\\
};
\addplot [color=blue, forget plot]
  table[row sep=crcr]{%
2.75	0.000180742663897782\\
2.75	0.0897542941407297\\
3.25	0.0897542941407297\\
3.25	0.000180742663897782\\
2.75	0.000180742663897782\\
};
\addplot [color=blue, forget plot]
  table[row sep=crcr]{%
3.75	0.029565185294512\\
3.75	0.102885040933049\\
4.25	0.102885040933049\\
4.25	0.029565185294512\\
3.75	0.029565185294512\\
};
\addplot [color=blue, forget plot]
  table[row sep=crcr]{%
4.75	0.00354994750730153\\
4.75	0.136397085417089\\
5.25	0.136397085417089\\
5.25	0.00354994750730153\\
4.75	0.00354994750730153\\
};
\addplot [color=blue, forget plot]
  table[row sep=crcr]{%
5.75	0.000399291080679376\\
5.75	0.173584241220238\\
6.25	0.173584241220238\\
6.25	0.000399291080679376\\
5.75	0.000399291080679376\\
};
\addplot [color=red, forget plot]
  table[row sep=crcr]{%
1.75	0.00976401558511153\\
2.25	0.00976401558511153\\
};
\addplot [color=red, forget plot]
  table[row sep=crcr]{%
2.75	0.0466933832128049\\
3.25	0.0466933832128049\\
};
\addplot [color=red, forget plot]
  table[row sep=crcr]{%
3.75	0.033979037845449\\
4.25	0.033979037845449\\
};
\addplot [color=red, forget plot]
  table[row sep=crcr]{%
4.75	0.0559010701042881\\
5.25	0.0559010701042881\\
};
\addplot [color=red, forget plot]
  table[row sep=crcr]{%
5.75	0.0576748623247836\\
6.25	0.0576748623247836\\
};
\addplot [color=black, draw=none, mark=+, mark options={solid, red}, forget plot]
  table[row sep=crcr]{%
2	0.128354017488427\\
2	0.169829627685652\\
2	0.178569986166916\\
2	0.179398615666192\\
};
\end{axis}

\begin{axis}[%
width=0.409\figurewidth,
height=\figureheight,
at={(0.48\figurewidth,0\figureheight)},
scale only axis,
xmin=1.5,
xmax=6.5,
xtick={1,2,3,4,5,6},
xticklabels={{0},{1.5},{2},{2.5},{3},{3.5}},
xlabel style={font=\color{white!15!black}},
xlabel={Mean depth measured [m]},
ymin=0,
ymax=0.3,
axis background/.style={fill=white},
title style={font=\bfseries},
title={y},
title style={font={\small\bfseries}},legend style={font=\tiny},
]
\addplot [color=black, dashed, forget plot]
  table[row sep=crcr]{%
2	0.0171988499606015\\
2	0.0410519862511127\\
};
\addplot [color=black, dashed, forget plot]
  table[row sep=crcr]{%
3	0.0210678978474892\\
3	0.0489905689185077\\
};
\addplot [color=black, dashed, forget plot]
  table[row sep=crcr]{%
4	0.0528922506270231\\
4	0.0667543586436494\\
};
\addplot [color=black, dashed, forget plot]
  table[row sep=crcr]{%
5	0.0567608096361756\\
5	0.127486744436738\\
};
\addplot [color=black, dashed, forget plot]
  table[row sep=crcr]{%
6	0.0516779952029154\\
6	0.124453934423303\\
};
\addplot [color=black, dashed, forget plot]
  table[row sep=crcr]{%
2	5.40244519164901e-06\\
2	0.000192358028083817\\
};
\addplot [color=black, dashed, forget plot]
  table[row sep=crcr]{%
3	2.64552800879336e-05\\
3	0.000330492571694485\\
};
\addplot [color=black, dashed, forget plot]
  table[row sep=crcr]{%
4	0.0159022570317058\\
4	0.0217011432694774\\
};
\addplot [color=black, dashed, forget plot]
  table[row sep=crcr]{%
5	0.000216904969798648\\
5	0.00627720205204874\\
};
\addplot [color=black, dashed, forget plot]
  table[row sep=crcr]{%
6	8.92577711919529e-06\\
6	0.000944523468431768\\
};
\addplot [color=black, forget plot]
  table[row sep=crcr]{%
1.875	0.0410519862511127\\
2.125	0.0410519862511127\\
};
\addplot [color=black, forget plot]
  table[row sep=crcr]{%
2.875	0.0489905689185077\\
3.125	0.0489905689185077\\
};
\addplot [color=black, forget plot]
  table[row sep=crcr]{%
3.875	0.0667543586436494\\
4.125	0.0667543586436494\\
};
\addplot [color=black, forget plot]
  table[row sep=crcr]{%
4.875	0.127486744436738\\
5.125	0.127486744436738\\
};
\addplot [color=black, forget plot]
  table[row sep=crcr]{%
5.875	0.124453934423303\\
6.125	0.124453934423303\\
};
\addplot [color=black, forget plot]
  table[row sep=crcr]{%
1.875	5.40244519164901e-06\\
2.125	5.40244519164901e-06\\
};
\addplot [color=black, forget plot]
  table[row sep=crcr]{%
2.875	2.64552800879336e-05\\
3.125	2.64552800879336e-05\\
};
\addplot [color=black, forget plot]
  table[row sep=crcr]{%
3.875	0.0159022570317058\\
4.125	0.0159022570317058\\
};
\addplot [color=black, forget plot]
  table[row sep=crcr]{%
4.875	0.000216904969798648\\
5.125	0.000216904969798648\\
};
\addplot [color=black, forget plot]
  table[row sep=crcr]{%
5.875	8.92577711919529e-06\\
6.125	8.92577711919529e-06\\
};
\addplot [color=blue, forget plot]
  table[row sep=crcr]{%
1.75	0.000192358028083817\\
1.75	0.0171988499606015\\
2.25	0.0171988499606015\\
2.25	0.000192358028083817\\
1.75	0.000192358028083817\\
};
\addplot [color=blue, forget plot]
  table[row sep=crcr]{%
2.75	0.000330492571694485\\
2.75	0.0210678978474892\\
3.25	0.0210678978474892\\
3.25	0.000330492571694485\\
2.75	0.000330492571694485\\
};
\addplot [color=blue, forget plot]
  table[row sep=crcr]{%
3.75	0.0217011432694774\\
3.75	0.0528922506270231\\
4.25	0.0528922506270231\\
4.25	0.0217011432694774\\
3.75	0.0217011432694774\\
};
\addplot [color=blue, forget plot]
  table[row sep=crcr]{%
4.75	0.00627720205204874\\
4.75	0.0567608096361756\\
5.25	0.0567608096361756\\
5.25	0.00627720205204874\\
4.75	0.00627720205204874\\
};
\addplot [color=blue, forget plot]
  table[row sep=crcr]{%
5.75	0.000944523468431768\\
5.75	0.0516779952029154\\
6.25	0.0516779952029154\\
6.25	0.000944523468431768\\
5.75	0.000944523468431768\\
};
\addplot [color=red, forget plot]
  table[row sep=crcr]{%
1.75	0.0026117108077246\\
2.25	0.0026117108077246\\
};
\addplot [color=red, forget plot]
  table[row sep=crcr]{%
2.75	0.00631358013369621\\
3.25	0.00631358013369621\\
};
\addplot [color=red, forget plot]
  table[row sep=crcr]{%
3.75	0.0381190423283115\\
4.25	0.0381190423283115\\
};
\addplot [color=red, forget plot]
  table[row sep=crcr]{%
4.75	0.0204916969028321\\
5.25	0.0204916969028321\\
};
\addplot [color=red, forget plot]
  table[row sep=crcr]{%
5.75	0.022303567973335\\
6.25	0.022303567973335\\
};
\addplot [color=black, draw=none, mark=+, mark options={solid, red}, forget plot]
  table[row sep=crcr]{%
6	0.134137385750505\\
};
\end{axis}
\end{tikzpicture}%

	\caption{Boxplot of the absolute velocity estimate error of Edge-FS, compared against the mean observed depth.} 
	\label{fig:matlab_windowsize_quality}
\end{figure}

It is important to note that because of the nonlinear relation between pixel disparity and depth in stereo vision, far distances are measured less accurately. The disparities for further distances will become sub-pixel and hard to determine. This is especially relevant to the small stereo board used in this study, which we set up to use 128 x 96 pixel images for the 57.4 x 44.5 deg FOV. Also, the translational optical flow of objects is harder to measure when they are further away, since it becomes sub-pixel as well. Hence, both terms on the left in (\ref{eq:linefit}), $\mathbf{s}_u$ and $\mathbf{d}_x$ become less accurate at far distances. This correlation between distance and accuracy can be seen in the box plot of Fig.~\ref{fig:matlab_windowsize_quality}. 

Besides the difficulty with larger distances, which is fundamental to stereo vision, Edge-FS also has a bit difficulty determining the forward flow when there is a large lateral motion. Sideways velocity has a $0^{th}$ order effect on the flow, while
forward velocity information is captured by the divergence of the flow field,
which is a $1^{st}$ order effect. Therefore, the forward velocity is more
subject to noise and harder to estimate (this can be observed in Fig.~\ref{fig:matlab_windowsize_quality} as the errors are generally higher for the x-direction than the y-direction). In this work, the \ac{MAV} will mostly fly forward. In this situation lateral flow is kept very small while the divergence is larger and more observable. A larger SAD window size and filtering are used to correct for the remaining noise.
While further analysis of the noise is beyond the scope of this article, the reader is encouraged to look at it in more detail by using our MATLAB code for Edge-FS including a large diverse image data-set\footnote{\url{https://github.com/knmcguire/EdgeFlow\textunderscore matlab}}.


\section{Experiments on the Pocket Drone}\label{sec:experiments}
In this section, we explain the implementation of Edge-FS on-board a pocket drone and how it is used in an autonomous obstacle avoidance task. We will first present the velocity estimates by Edge-FS during flight on the stereo board. Subsequently, a closed loop flight is shown, where the drone autonomously navigates through a room, while maintaining its velocity and avoiding obstacles.
\subsection{Hardware specifics}

\begin{figure}[t]
	\vspace{2pt}
	\centering
	
	\scriptsize	\def\svgwidth{0.55\linewidth}
	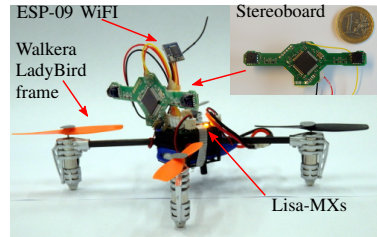
	\vspace{-0.3cm}
	\caption{The 4~g stereo camera mounted on the pocket drone.
		 \vspace{-0.3cm}} 
	\label{fig:setup_ardrone_stereocamera}
\end{figure}

Edge-FS local runs embedded on the stereo camera (as introduced in~\cite{de2014autonomous}).  Fig.~\ref{fig:setup_ardrone_stereocamera} displays two cameras with 1/6 inch image sensors, with a baseline of 6 cm and a \ac{FOV} of $57.5^o$ x $44.5^o$. The stereo camera has an embedded microprocessor, an STM32F4 with a speed of 168 MHz and 196 kB of memory in which the largest consecutive memory block spans 128 kB. The cameras are configured to output stereo-images with a size of 128 x 96 pixels to fit within memory and processing constraints. The maximum reachable frame rate of the stereo camera is 30 Hz, which is not much affected by the computation of Edge-FS (approx. 0.0175 sec).

For the experiments, a pocket drone is equipped with a single front-facing stereo camera. A frame of a  Walkera QR LadyBug\footnote{\url{http://www.walkera.com/}} is adopted as a base. An adapted smaller variant of the Lisa-MX\footnote{\url{http://wiki.paparazziuav.org/wiki/Lisa/MX}} will be used the auto-pilot. The \textit{Lisa-MXs} also carries an STM32F4 microprocessor, with a speed of 168 Hz and 1 MB of flash memory. With an ESP-09 WiFi module, telemetry can be broadcasted to the computer to receive all the measured variables required for validation. The entire assembly, including stereo camera and battery, weighs exactly 41.9 g.

The auto-pilot program flashed on the Lisa-MXs is Paparazzi\footnote{\url{http://wiki.paparazziuav.org/}}. The software runs entirely on-board the microprocessor which governs all the basic flight controls. An adaptive Incremental Nonlinear Dynamic Inversion (INDI) controller~\cite{smeur2015adaptive} is used for the attitude stabilization of the \ac{MAV}. The guidance controller resides on top of the stabilization control, to calculate the desired pitch and roll angle, to achieve a desired altitude position or airspeed. In this paper, it will be applied to maintain a desired velocity. It will need the measurements from the stereo camera, operating in parallel with the Lisa-MXs.

\subsection{Velocity Estimate}\label{sec:velocityest}

We have shown in section~\ref{sec:simulation} that EdgeFlow can measure the camera velocity based on a collection of images. Now implemented in the 4~g stereoboard and fixed on a pocket drone, the question remains if it can still retain its quality with all the additional effects caused by motion and vibrations during flight.

\begin{figure}[t]
\centering
\small
\subfigure[\hspace{4cm}\label{subfig:vel_est_ATT}]{\hspace{-1cm}\setlength\figureheight{5cm} 
\setlength\figurewidth{0.5\linewidth}
\input{\matlabtikz/Edgeflow_velocity_estimate_ATT}}\hspace{-0.5cm}
\begin{minipage}{0.33\linewidth}
	\vspace{-6.0cm}
\addtolength{\subfigcapskip}{+10pt}
	\subfigure[\hspace{1.5cm}\label{subfig:vel_est_screenshot_ATT}]{		\hspace{0.9cm}	\includegraphics[ width  = 0.8\linewidth]{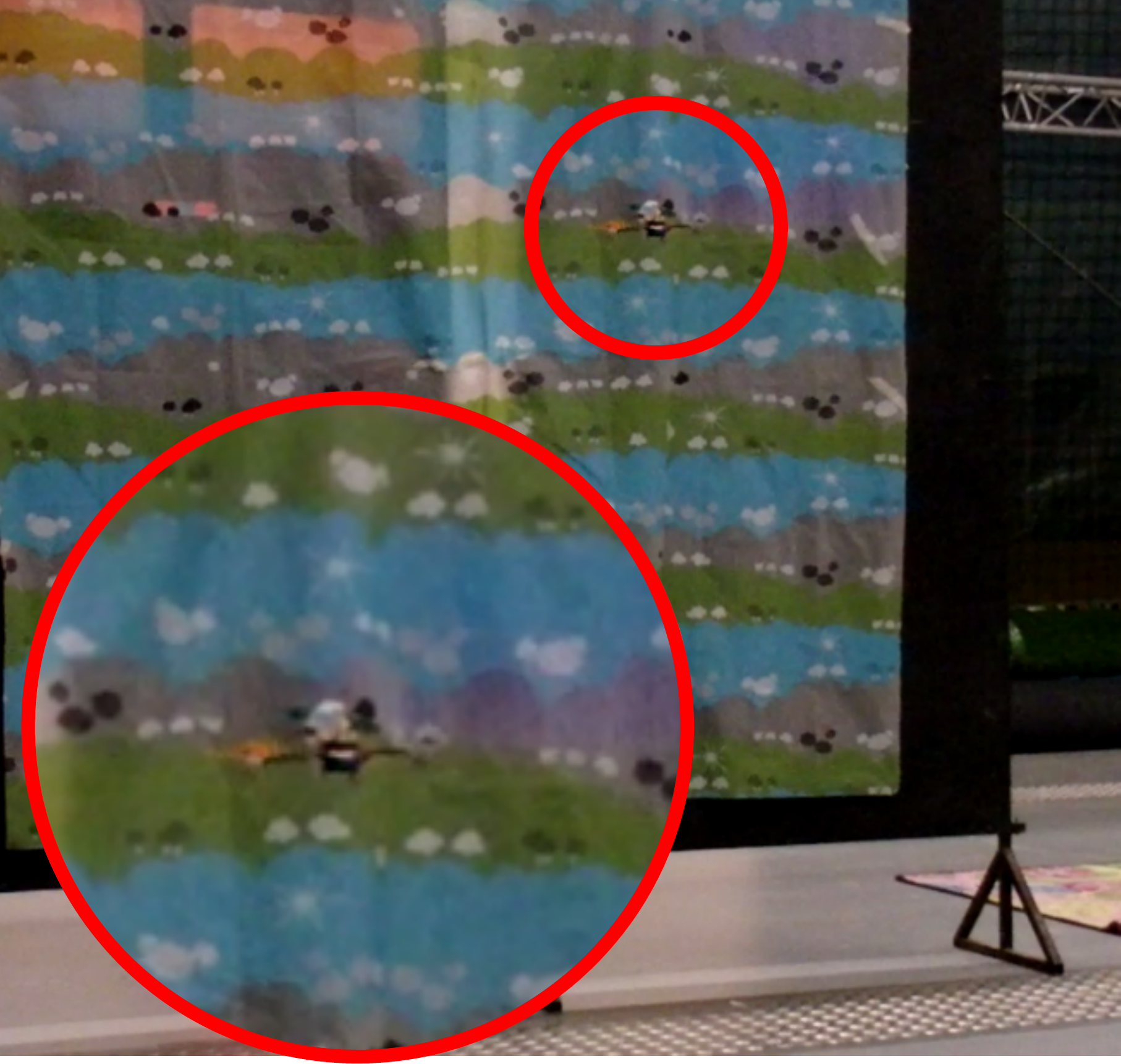} }
	\vspace{-0.3cm}	\addtolength{\subfigcapskip}{-10pt}
\subfigure[\hspace{1.5cm}\label{subfig:vel_est_pos_ATT}]{		\setlength\figureheight{0.8\linewidth} 

		\setlength\figurewidth{0.8\linewidth}
		\input{\matlabtikz/Edgeflow_position_ATT}
	} 
\end{minipage}

\caption{(a) Velocity estimation by Edge-FS on the pocket drone, (b) a screen shot from the flight in front of a wall and (c) the position during a remote controlled maneuver. Dotted line is the unfiltered velocity estimate by Edge-FS.} 
\label{fig:velocity_estimate}
\end{figure}

Fig.~\ref{subfig:vel_est_ATT}, presents the velocity estimates of Edge-FS, during a manually controlled flight in front of a textured screen (screen shot in Fig.~\ref{subfig:vel_est_screenshot_ATT} and position in Fig.~\ref{subfig:vel_est_pos_ATT} ). The same OptiTrack system used for the image data set (Fig.~\ref{subfig:database_forward}), is monitoring its real velocity. The raw unfiltered velocity measurements of Edge-FS are contrasted with this ground truth with \ac{NMXM}, \ac{VAR} and \ac{MSE}. Noticeable is that the forward velocity shows more noise peaks than the sideways velocity as expected (see Section~\ref{sec:simulation}). However, in both directions, Edge-FS matches the ground truth adequately, which should be sufficient for the closed-loop flight.

To use the actual raw measurements in flight is undesirable. The most common way is to fuse these vision-based velocity estimates with the accelerometers. On a larger \ac{MAV} than the pocket drone, this would be possible because of the damping. However, many vibrations are generated by the small propellers, which are in close proximity with the autopilot, the accelerometers readings contain too much noise.
Therefore, in this paper, we only use a vision-only approach applying a median filter to the 5 last velocity measurements, to keep the delay to a minimum.

\begin{figure}[t]
	
\vspace{-1cm}
\hspace{1cm}
	\flushright
	\small
	
	\subfigure[\hspace{4cm}\label{subfig:velocitycontrol_plot}]{\hspace{-0.3cm}\setlength\figureheight{5cm} 
		\setlength\figurewidth{0.5\linewidth}

\vspace{1cm}
		\input{\matlabtikz/Edgeflow_velocity_guided}}\subfigure[\hspace{4cm}\label{subfig:control_scheme2}]{	
		\hspace{-0.3cm}
			\scriptsize	\def\svgwidth{0.4\linewidth}
	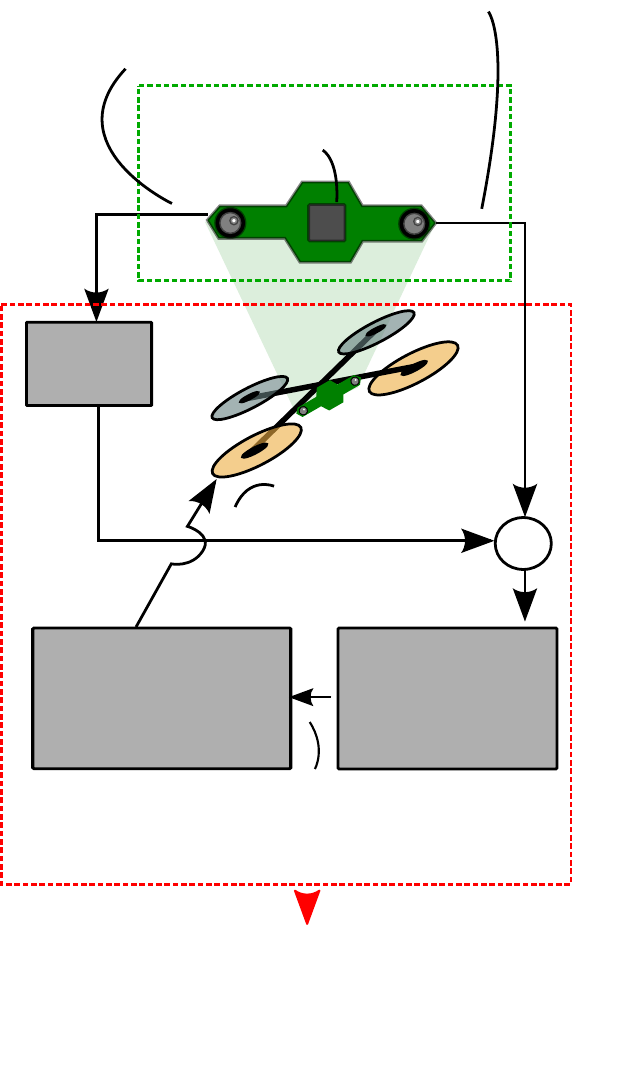

	}

	\vspace{-0.3cm}
	\caption{(a) Velocity control on pocket drone with a wall force field and (b) the control scheme used, explaining the hierarchy of the on-board sensors and controllers. } 
	\label{fig:velocity_control}
\end{figure}

\subsection{Autonomous Obstacle Avoidance} \label{sec:closedloopflight}

\begin{figure}[t]
	\vspace{0.5cm}
	\flushright
	\scriptsize 
	\vspace{-0.5cm}
	\addtolength{\subfigcapskip}{+5pt}
\subfigure[\vspace{-0.3cm}\hspace{4cm}\label{subfig:closedloopflight_plot}]{\setlength\figureheight{0.45\linewidth} 
	\setlength\figurewidth{0.45\linewidth}
	\input{\matlabtikz/Edgeflow_position_CL}}	\subfigure[
\hspace{4cm}\label{subfig:avoidancelogic}]{	
	\scriptsize	\def\svgwidth{0.4\linewidth}
	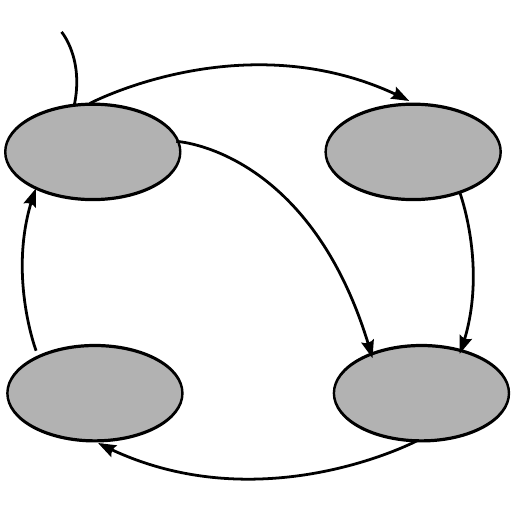}
\centering
\vspace{-0.2cm}

\subfigure[\vspace{-0.3cm}\hspace{0cm}\label{subfig:closedloopflight_screenshot}]{
		\vspace{-0.3cm}
	\includegraphics[ width  = 0.3\linewidth]{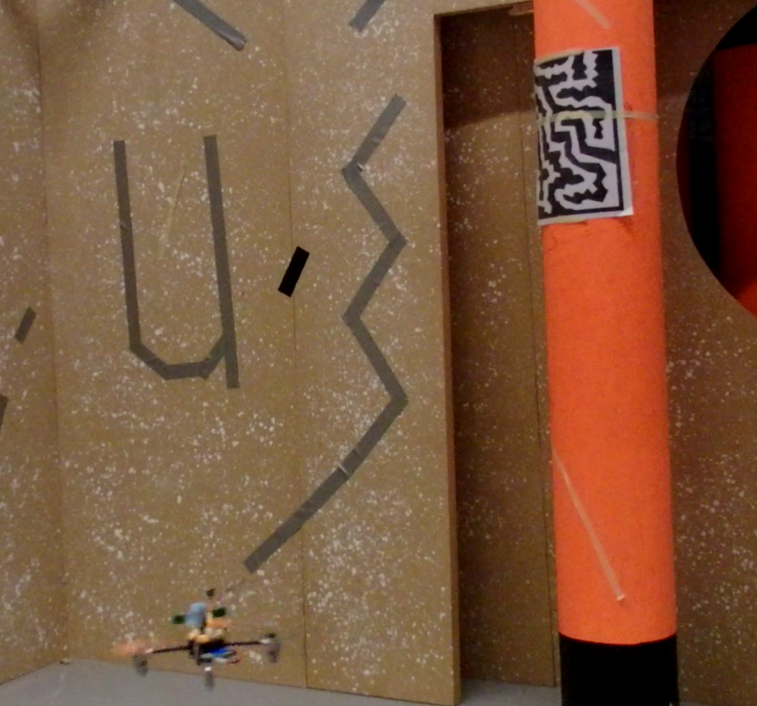} \hspace{1cm} \includegraphics[ width  = 0.3\linewidth]{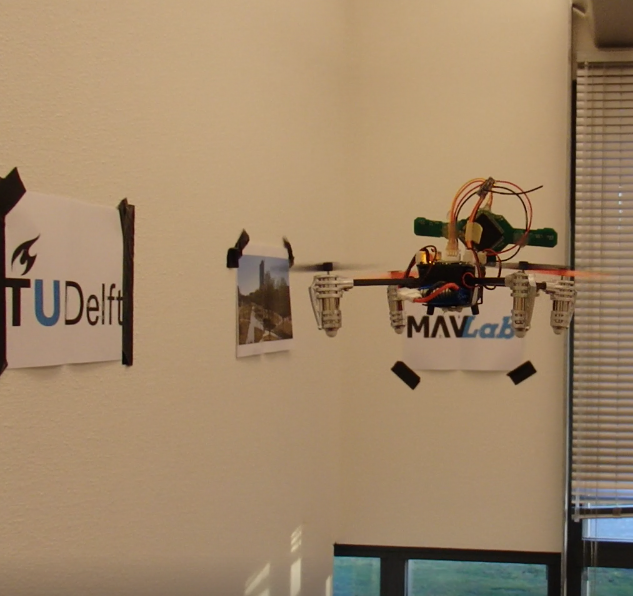}}
\textsl{}
	
	\caption{(a) A position plot of 3 flights, from which the first lasted 91 seconds, the second 101 seconds and the third 122 seconds. (b) shows screen shots of the experiments in the flight arena (left) and in a real-world office (right). Some posters were added to the latter to provide extra texture.}
	\label{fig:closed_loop_flight}
\end{figure}

In the previous subsection, we showed validation of the velocity estimate as calculated by Edge-FS. Now we will present a closed-loop flight, where the pocket drone avoids obstacles identified by means of its stereo vision, while guided by its velocity estimates. The main goal of this experiment is to show the potential of the proposed algorithms for full autonomous navigation. In this section, the vertical position as measured by OptiTrack is exclusively used for height control, as no position measurement is used in the horizontal plane (solely for validation afterwards). This is where the \ac{MAV} uses its velocity estimates by Edge-FS.

Fig.~\ref{subfig:control_scheme2} displays the basic control scheme for the navigation task. It determines a desired velocity to avoid collisions. The error between the estimate and the desired velocity is the input to the velocity guidance controller, which sets an attitude set-point for the stabilization. Subsequently, EdgeStereo determines the nearest object to camera. If too close, it will produce a backward velocity reference to the guidance controller (a force field), therefore preventing the pocket drone from hitting the wall face-on. Fig.~\ref{subfig:velocitycontrol_plot} shows the readings from a short flight of a simple hover with the obstacle force field.

When encountering a wall/obstacle, the pocket drone will need to move away from the situation. The avoidance scheme is a simple finite state machine (FSM) with 4 behavioral states (see Fig.~\ref{subfig:avoidancelogic}). It starts in \textit{check} mode, where the pocket drone will check if there is a detected obstacle within 1 meter by EdgeStereo. If the way is clear, the pocket drone moves \textit{forward} with a constant speed (set now to 0.3 m/s), guided by the velocity estimate from Edge-FS. If it detects an object on its path, the \ac{MAV} will first hover for 1 second actively controlling the forward velocity to zero. Then it will turn quickly with a constant angle relative to the heading (here $\Delta\psi = 60^o$). Immediately thereafter, the \ac{MAV} will evaluate the situation in the \textit{check} mode and proceeds from there.
 
We conducted multiple autonomous flights with the pocket drone. Fig~\ref{subfig:closedloopflight_plot} shows the result of 3 representative flights of the pocket drone with the forward looking stereo camera. The pocket drone has to navigate in a small room of 4 x 4 meter with varying textured surfaces (screen shot of camera footage). All the flights lasted longer than 90 seconds, from which the longest duration was 122 seconds (flight 3). When the pocket drone brushed against the wall, the safety pilot took over the flight with a remote control for a safe landing. The most common failure case during the test flights, is that the \ac{MAV} will approach the wall with a small angle. After the turn with constant angle, the drone will fly almost parallel to the wall which it can not detect due to its limited \ac{FOV}. This is the case for flight 2 and 3, except for flight 1, in which case the pocket drone was facing the observer after a turn. Several flights of the pocket drone have been done within a real-world environment (Fig.~\ref{subfig:closedloopflight_screenshot}), which can be observed with the accompanying video and YouTube list\footnote{\scriptsize\url{https://www.youtube.com/playlist?list=PL\textunderscore KSX9GOn2P812tmddfrTlURHNieRe6YY}} in Fig.~\ref{subfig:closedloopflight_screenshot}.

The mentioned failure case for the autonomous flights is difficult to overcome. If the \ac{MAV} would turn and face a large open space, the distance for EdgeStereo could be far enough to compromise the quality for the velocity estimate due to the small base line of the stereo camera. As we already observed in Fig.~\ref{fig:matlab_windowsize_quality}, this would cause the pocket drone to drift, which is problematic when near a wall/obstacle after the turn. If an obstacle is not in its FOV, the chances of collision significantly increases. This could be solved by merging the \textit{check} and \textit{turn} node of the FSM, so it will only stop turning at a significant clear path. Another solution is to add a lightweight short range sensor on the sides of the pocket drone, so it will detect immediately if the drone is flying close and aside an obstacle.

The obstacle avoidance logic will need some additional work, however the experiments show that Edge-FS can be used in navigation overall. During the autonomous flight, the pocket drone was stabilizing itself using the velocity estimates of its forward camera alone.

\section{Conclusion}
A computationally  efficient optical flow and stereo algorithm is presented in this paper, called Edge-FS. It runs embedded on a very lightweight stereo camera and can be carried by a 40~g pocket drone for determining velocity and depth. The presented algorithm allows the stereo camera to face forward, a direction in which a complex 3D structure can be expected.

We presented experiments where the pocket drone with the stereo camera autonomously navigated and avoided obstacles in an area of 4 x 4 meters. A simple finite state machine controller showed that the velocity estimates and the depth measurement can be used for fully autonomous flight. The current work lays the basis for stabilization and collision avoidance on pocket drones with a single, small stereo vision system.

\section*{Acknowledgments}
We would like to show our gratitude to Kirk Scheper from the MAVlab, Delft University of Technology, for assistance in the development of the guided flight mode and the finite state machine. These results greatly improved the quality of the paper.

\bibliographystyle{IEEEtran}
\bibliography{IEEEabrv,library_RAS_journal}
\end{document}